\documentclass[letterpaper, 10 pt, conference]{ieeeconf}  % Comment this line out if you need a4paper

\IEEEoverridecommandlockouts                              % This command is only needed if 
\usepackage{soul}         % 用于文本荧光标记
\sethlcolor{yellow}       % 设置高亮颜色为黄色
\usepackage{graphics} % for pdf, bitmapped graphics files
\usepackage{epsfig} % for postscript graphics files
\usepackage{mathptmx} % assumes new font selection scheme installed
\usepackage{times} % assumes new font selection scheme installed
\usepackage{amsmath} % assumes amsmath package installed
\usepackage{amssymb}  % assumes amsmath package installed
\usepackage{booktabs}
\usepackage{xcolor}

\usepackage{booktabs}       % Professional-quality tables
\usepackage{amsfonts}       % Blackboard math symbols
\usepackage{nicefrac}       % Compact symbols for 1/2, etc.
\usepackage{microtype}      % Microtypography
\usepackage{xcolor}         % Colors
\usepackage{graphicx}       % Provides \resizebox
\usepackage{amsmath}
\usepackage{amssymb}
\usepackage{colortbl}       % Cell background colors
\usepackage{pifont} 
\newcommand{\cmark}{\ding{51}} % Checkmark
\newcommand{\xmark}{\ding{55}} % Cross
\definecolor{waymogreen}{HTML}{00E89D}
\definecolor{waymolgreen}{HTML}{99F7D7} % alpha=60%
\definecolor{waymollgreen}{HTML}{CCFAEB} % alpha=80%
\definecolor{waymoblue}{HTML}{0077FF}
\definecolor{waymolblue}{HTML}{99B7FF}  % alpha=60%
\definecolor{waymollblue}{HTML}{CCE4FF} % alpha=80%
\definecolor{waymolgray}{HTML}{F0F0F0}  % 浅灰色

\usepackage[colorlinks=false,      % false: boxed links; true: colored links
linkcolor=black,      % color of internal links
citecolor=black,      % color of links to bibliography
filecolor=black,      % color of file links
urlcolor=blue]{hyperref}

\title{\LARGE \bf
FASIONAD++ \raisebox{-0.1\height}{\includegraphics[width=0.04\linewidth]{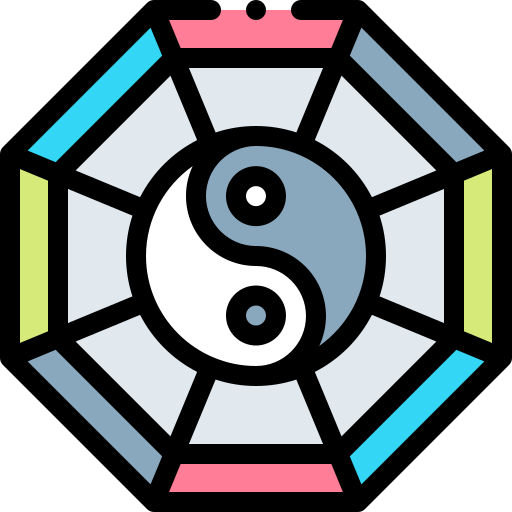}} : 
Integrating High-Level Instruction and Information Bottleneck in FASt-Slow fusION Systems for Enhanced Safety in Autonomous Driving with Adaptive Feedback}

\author{Kangan Qian$^{1, \dagger}$, Ziang Luo$^{1, \dagger}$, Sicong Jiang$^{2}$, Zilin Huang$^{3}$, Jinyu Miao$^{1}$, Zhikun Ma$^{4}$, Tianze Zhu$^{1}$, \\ 
Jiayin Li $^{5}$, 
Yangfan He$^{5}$, Zheng Fu$^{1}$, Yining Shi$^{1}$, Boyue Wang$^{3}$, Hezhe Lin$^{1}$, \\ Ziyu Chen$^{6,*}$, Jiangbo Yu$^{2}$, 
Xinyu Jiao$^{1}$, 
Mengmeng Yang$^{1}$, Kun Jiang$^{1,*}$, Diange Yang$^{1,*}$
\thanks{*This work was not supported by any organization.}% <-this % stops a space
\thanks{$^{\dagger}$The authors contribute equally to this work.}
\thanks{$^{1}$The School of Vehicle and Mobility, Tsinghua University, Beijing, China. {\tt\small qka23@mails.tsinghua.edu.cn};
$^{2}$McGill University, Canada; $^{3}$University of Wisconsin-Madison, USA; $^{4}$Waseda University, Japan; $^{5}$University of Minnesota, USA;  $^{6}$AI2Robotics, Beijing, China.}%
\thanks{Kangan Qian was with AI2Robotics during his internship in Beijing, China.}
\thanks{$^{*}$Corresponding author: Ziyu Chen, Kun Jiang, and Diange Yang.}
}

\begin{document}

\maketitle
\thispagestyle{empty}
\pagestyle{empty}

%%%%%%%%%%%%%%%%%%%%%%%%%%%%%%%%%%%%%%%%%%%%%%%%%%%%%%%%%%%%%%%%%%%%%%%%%%%%%%%%
\begin{abstract}
Ensuring safe, comfortable, and efficient planning is crucial for autonomous driving systems. While end-to-end models trained on large datasets perform well in standard driving scenarios, they struggle with complex low-frequency events. Recent Large Language Models (LLMs) and Vision Language Models (VLMs) advancements offer enhanced reasoning but suffer from computational inefficiency. Inspired by the dual-process cognitive model  \emph{``Thinking, Fast and Slow''}, we propose \textbf{FASIONAD} -- a novel dual-system framework that synergizes a fast end-to-end planner with a VLM-based reasoning module. The fast system leverages end-to-end learning to achieve real-time trajectory generation in common scenarios, while the slow system activates through uncertainty estimation to perform contextual analysis and complex scenario resolution. Our architecture introduces three key innovations: (1) A dynamic switching mechanism enabling slow system intervention based on real-time uncertainty assessment; (2) An information bottleneck with high-level plan feedback that optimizes the slow system's guidance capability; (3) A bidirectional knowledge exchange where visual prompts enhance the slow system's reasoning while its feedback refines the fast planner's decision-making. To strengthen VLM reasoning, we develop a question-answering mechanism coupled with reward-instruct training strategy. In open-loop experiments, FASIONAD achieves a $6.7\%$ reduction in average $L2$ trajectory error and $28.1\%$ lower collision rate. 

\end{abstract}

%%%%%%%%%%%%%%%%%%%%%%%%%%%%%%%%%%%%%%%%%%%%%%%%%%%%%%%%%%%%%%%%%%%%%%%%%%%%%%%%
\section{INTRODUCTION}

\begin{figure*}[ht]
    \centering
    \includegraphics[scale=0.5]{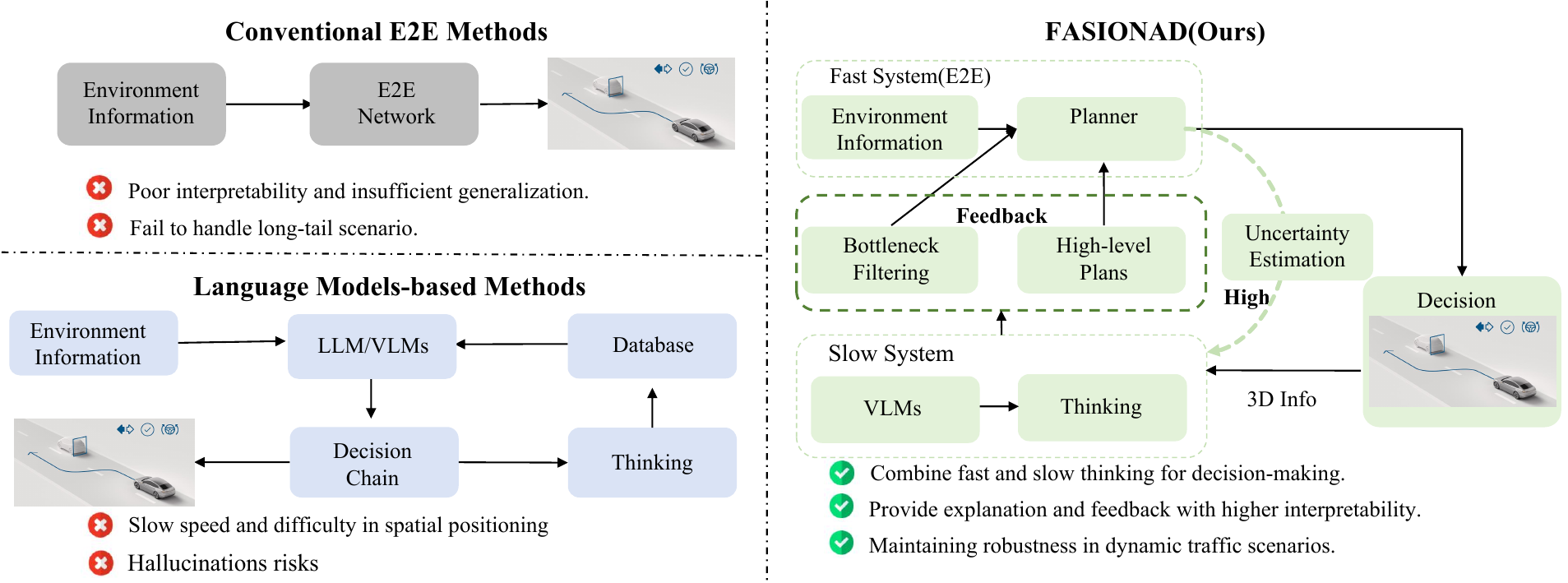}
    \caption{
   The motivation of our FASIONAD. Conventional E2E methods struggle with interpretability and generalization. LLMs-based methods face slow decision-making, spatial positioning issues, and potential hallucinations. We compares different motion planning methods for autonomous driving, showcasing our method’s ability to adaptive, context-aware decisions, offering better explanation and feedback. 
    }
    \label{fig:motivation}
\end{figure*}

As technology advances, autonomous driving holds the potential to transform transportation by enhancing efficiency, reducing human workload, and minimizing accidents~\cite{kiran2021deep}. Traditional autonomous driving systems typically follow a modular design consisting of perception, prediction, and planning~\cite{kiran2021deep, zhou2022dynamically, shi2019driving}. Although such modular approaches provide interpretability, they can be rigid in nature (\emph{e.g.}, rule-based controllers) and may struggle to handle complex, dynamic real-world scenarios~\cite{zhou2022dynamically}. In contrast, End-to-End (E2E) learning methods have recently gained attention, aiming to learn driving policies directly from sensory inputs~\cite{ jiang2024communication, jiang2023vad}. However, purely E2E models often exhibit insufficient generalization and reliability, especially in long-tail driving situations~\cite{liu2024curse}. Attempts to refine E2E systems with a trajectory evaluation module often rely on open-loop evaluations (selecting trajectories without real-time feedback), making them susceptible to unforeseen failures.

Building on the surge of Large Language Models (LLMs) and Vision-Language Models (VLMs), researchers have recently explored how these models can aid autonomous driving tasks such as multi-modal perception~\cite{shao2024lmdrive} and high-level reasoning~\cite{wang2024safe}. Pure Language Model-based methods, however, face significant challenges with computational efficiency, reliability, and the high cost of training or fine-tuning~\cite{manipulation}. To address these issues, some works adopt dual-process  paradigms in an asynchronous manner~\cite{tian2024drivevlm}, but they do not explicitly switch between dual-process based on different driving contexts. As a result, these systems do not selectively leverage VLMs when complex reasoning is necessary, thus incurring excessive computation and introducing latency in scenarios where straightforward decisions suffice.

In practice, human drivers engage in in-depth reasoning only under specific circumstances. Most driving tasks (\emph{e.g.}, lane-keeping or car-following) are relatively routine and do not require continuous high-level cognition. By reserving complex reasoning for critical moments rather than every instant, human drivers effectively balance efficiency and safety. This observation naturally motivates the following question: \textit{Is it possible to design a system that unifies the strengths of both E2E models and VLMs, enabling more effective driving by emulating human-like integration of diverse information and nuanced decision-making?}

Inspired by this observation, we propose \textbf{FASIONAD}---a unified framework that harnesses an E2E (\emph{fast system}) policy for common driving tasks and a VLM (\emph{slow system}) for high-uncertainty or high-risk scenarios. As shown in Fig. \ref{fig:motivation}, FASIONAD employs an adaptive switching mechanism to activate the slow system only when additional reasoning depth is required, effectively mitigating computational overhead. Unlike prior methods that rely on direct VLM trajectory outputs, our framework leverages the VLM for feedback and evaluation through concise, deterministic cues. We posit that many E2E failures arise from decision-making rather than perception, and thus introduce information bottleneck filtering to refine planning-oriented features and high-level action guidance to integrate VLM insights at a strategic level. Additionally, we incorporate precise Bird’s-Eye-View (BEV) and visual prompts from the fast system’s perception module to reduce uncertainty for VLM outputs.

We validate \textbf{FASIONAD} on the nuScenes~\cite{nuscenes_benchmark}, Town05 Short~\cite{carla_benchmark}, and Bench2Drive~\cite{b2d} benchmarks, where extensive experiments confirm the framework’s effectiveness in both routine and challenging scenarios. Our main contributions include:
\begin{itemize} 
\item Introducing \textbf{FASIONAD}, a dual-system autonomous driving framework that adaptively combines an E2E fast system with a VLM slow system for feedback-driven decision-making. 
\item Proposing three key modules—\textbf{Uncertainty Estimation} (UE) for adaptive switching, an \textbf{Information Bottleneck} (IB) filter to refine VLM inputs, and \textbf{High-level Action} (HA) guidance for strategic planning—that collectively enable targeted, interpretable feedback between the fast and slow systems.
\item Designing planning-oriented QAs using \textbf{visual} and \textbf{BEV prompts} to reduce VLM's unreliability. Empirical results show that our proposed FASIONAD significantly improves safety metrics with lower collision rate on the nuScenes, Town05 Short, and Bench2Drive benchmarks, across different fast system base models.\end{itemize}

\section{Related Work}

\subsection{Learning-based Planning} Navigating dynamic and complex environments is a key challenge in autonomous driving. Early methods typically employ modular pipelines for perception, planning, and control~\cite{kiran2021deep}, which offer interpretability but may hinder efficient information sharing among modules~\cite{yuan2024rag}. End-to-end (E2E) approaches, on the other hand, learn direct mappings from sensory inputs to control signals~\cite{codevilla2018end}, showing promising performance under routine conditions. Recent work extends E2E methods with Bird’s-Eye-View (BEV) representations to handle complex urban contexts~\cite{jiang2023vad, zheng2024genad}, aiming to improve spatial awareness and decision-making.
Despite these advances, purely E2E systems still suffer from limited interpretability and vulnerability to distributional shifts~\cite{chen2020learning}. Transformer-based solutions such as TransFuser~\cite{chitta2022transfuser} and InterFuser~\cite{shao2023safety} have introduced multi-modal fusion and attention mechanisms, achieving more robust predictions in diverse traffic scenarios. Yet, balancing real-time performance with high-level reasoning remains an active area of research.

\subsection{Vision-Language Models for Autonomous Driving} Vision-Language Models (VLMs) align visual and textual modalities to offer richer scene understanding~\cite{alayrac2022flamingo}. Foundational models such as CLIP and Flamingo~\cite{radford2021learning} demonstrate the potential for nuanced semantic representations, as showcased by Video-LLaVA~\cite{lin2023videollava} and DrivingCLIP~\cite{li2023drivingclip}, which support more detailed interpretations of dynamic driving scenarios. Beyond perception, VLMs have also been applied to high-level reasoning and planning tasks in multi-agent contexts, enhancing robustness in environments with complex interactions~\cite{fang2023video}.
Building on these insights, our proposed \textbf{FASIONAD} framework harnesses a VLM not just for semantic extraction but also for feedback-driven decision refinement in uncertain or rare scenarios. By selectively engaging the VLM, we address common E2E pitfalls—such as hallucinations and poor generalization—while preserving computational efficiency for routine driving tasks.

\begin{figure*}[ht]
    \centering
    \vspace{5pt}
    \includegraphics[width=\textwidth]{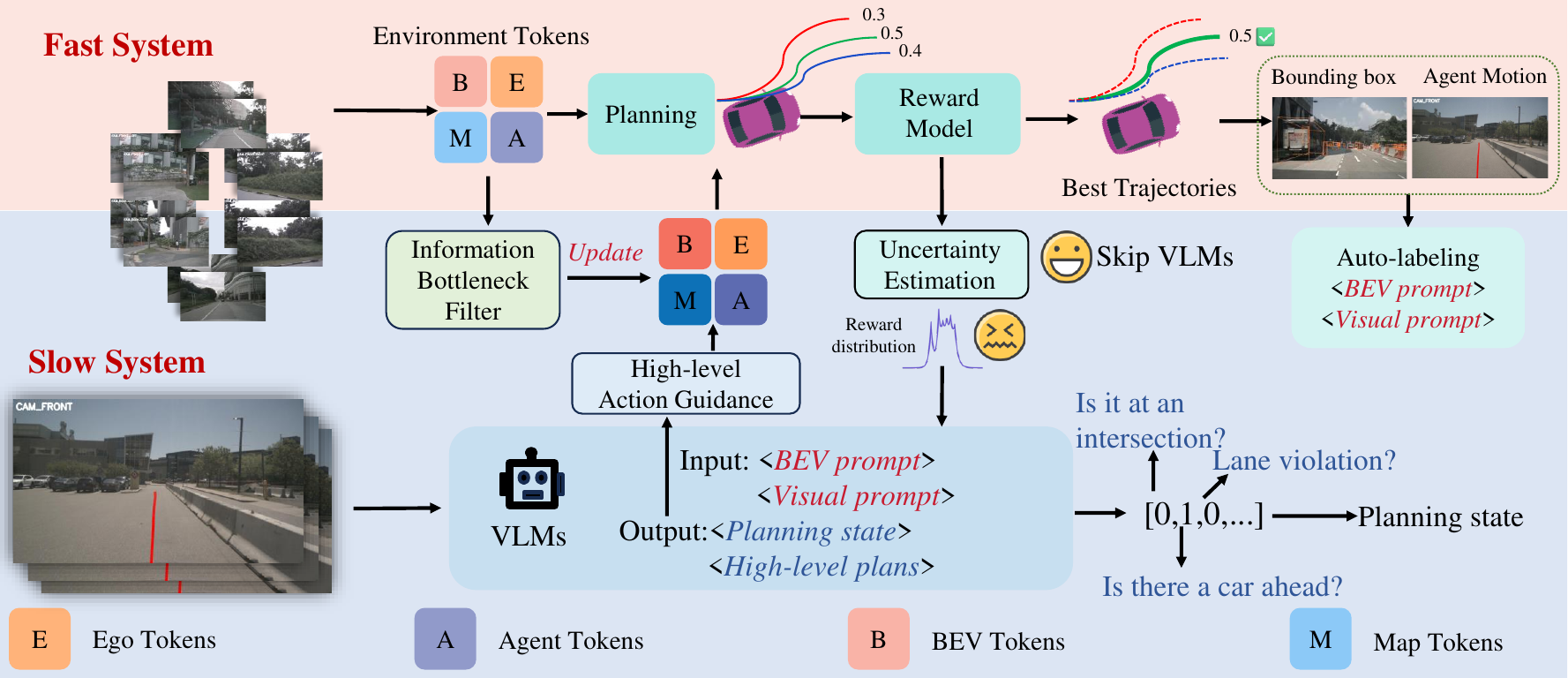}
    % \vspace{-10pt}
    \caption{
    The framework operates through dual-system: fast and slow. The fast ststem encodes image information into instance tokens(E, B, M, A relatively denotes ego tokens, BEV tokens, map tokens and agent tokens), generating multi-modal trajectories via a planning head. A reward model selects the optimal trajectory, while uncertainty estimation determines slow system activation. When engaged, the slow system utilizes VLM feedback, which is integrated both as HA and as scene-derived planning state vectors by IB, enabling trajectory refinement through the planning head.
    }
    \label{fig:framework}
    \vspace{-10pt}
\end{figure*}

\section{Methodology}
\subsection{Overview}
The inception of our methodology stems from the conviction that the primary challenge in E2E frameworks is not rooted in the precision of perception, but rather in aligning perception more closely with downstream planning, thereby genuinely embodying a planning-oriented paradigm. 

As depicted in Fig.~\ref{fig:framework}, FASIONAD employs a dual-system architecture: a fast system for rapid, real-time responses, and a slow system for comprehensive analysis and complex decision-making in uncertain or challenging driving scenarios. The fast system encodes image information into tokens, generating multi-modal trajectories along with a reward for each trajectory (Section \ref{fast}). In contrast, the slow system(Section \ref{slow}) processes the BEV prompt and visual prompt, subsequently outputting planning states and high-level plans for the entire driving scenario.

To ensure smooth coordination between the fast and slow systems, we have developed an innovative switching mechanism based on uncertainty estimation. This mechanism allows the fast system to refine its trajectory predictions by utilizing an information bottleneck and high-level plans derived from the slow system (Section \ref{fusion}).
\subsection{Fast System}\label{fast}
\textbf{Waypoints Prediction.}
Given a set of \( N \) multi-view images \( \textbf{I}_t = \{ I^1_{t}, I^2_{t}, \dots, I^N_{t} \} \) and high-level navigation commands \( \textbf{C}_t \), the model generates a sequence of waypoints \( \textbf{W}_t = \{ w^1_{t}, w^2_{t}, \dots, w^M_{t} \} \), where each waypoint \( w^i_{t} = [x^i_{t}, y^i_{t}] \) represents the predicted BEV position of the ego vehicle at time \( t + i \). This system can be formulated as:
\begin{equation}
\text{FASIONAD (fast system):} \quad (\textbf{I}_t, \textbf{C}_t) \rightarrow \textbf{W}_t.
\end{equation}

\textbf{Reward Evaluation.}
The model generates \( N_C \times N_{K} \) candidate trajectories \( \textbf{T} = \{{T}_i\}_{i=1}^{N_T} \), where each trajectory \( {T}_i \in \mathbb{R}^{\text{bs} \times T_s \times 2} \) represents a sequence of waypoints over a time horizon \( T_s \). Here, \( N_C \) is the number of navigation commands, and \( N_{K} \) represents the top-\( {K} \) sampled multi-modal trajectories. Each trajectory \( {T}_i \) is assigned a reward \( r_i \) by the reward model \( \mathcal{F}_{\text{Reward}} \), which integrates factors such as safety, comfort, efficiency, and economic considerations:
\begin{align}
    \mathcal{F}_{\text{Reward}} = &\ \alpha_{\text{safety}} C_{\text{safety}} 
    + \alpha_{\text{comfort}} C_{\text{comfort}} \notag \\
    &+ \alpha_{\text{efficiency}} C_{\text{efficiency}} 
    + \alpha_{\text{economic}} C_{\text{economic}}
\end{align}
where \( \alpha_{\text{safety}}, \alpha_{\text{comfort}}, \alpha_{\text{efficiency}}, \alpha_{\text{economic}} \) are weights determining the relative importance of each factor.

\subsection{Slow System}\label{slow}
\begin{figure*}[htbp]
    \centering
    % \vspace{-10pt}
    \includegraphics[width=\textwidth]{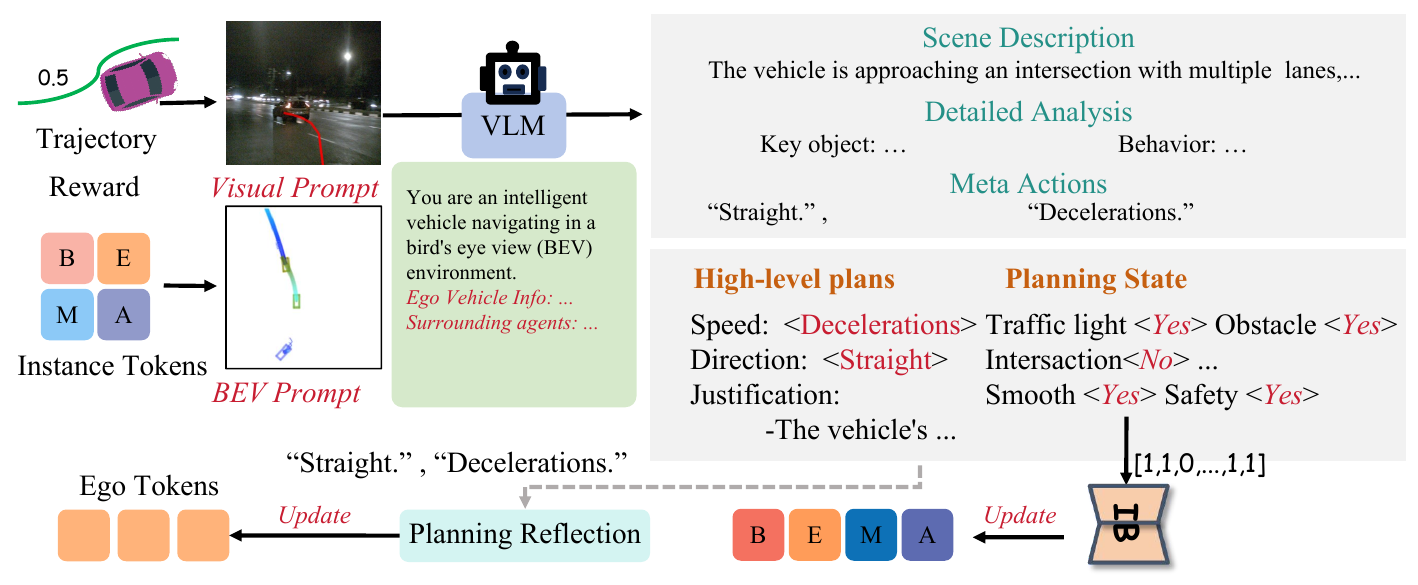}
    \vspace{-20pt}
    \caption{
    The adaptive feedback mechanism integrates dual inputs - visual prompts and BEV prompts - into a VLM. This VLM produces three outputs: scene descriptions, detailed analyses, and high-level plans, along with planning state vectors that encapsulate scene conditions. High-level plans are embedded into ego tokens, whereas planning state vectors pass through an IB to refine environment information in query tokens.
    }
    \label{fig:vlm}
    \vspace{-10pt}
\end{figure*}

In complex scenarios, accurate interpretation of environmental factors is vital for safe decision-making. The slow system emulates human-like reasoning to infer context and predict future actions, similar to human drivers. This section discusses how VLMs can support such reasoning, with a focus on QA design in Section \ref{sec:QA}, which formats the output of VLM models, and VLM tuning in Section \ref{sec:vlm-tunning}.

\subsubsection{Planning-oriented QA} \label{sec:QA} Building on existing QA frameworks for autonomous driving~\cite{tian2024drivevlm, jiang2024senna}, we propose a structured approach aimed at human-like reasoning. As illustrated in Fig.~\ref{fig:vlm}, our design centers on five key aspects critical to robust driving policies: \begin{itemize} \item[(i)] \textbf{Scene analysis:} Evaluates environmental conditions (e.g., weather, lighting, traffic density) to guide overall decision-making. \item[(ii)] \textbf{Traffic sign recognition:} Detects and interprets traffic signs for regulatory compliance. \item[(iii)] \textbf{Key object recognition:} Identifies and predicts the behavior of nearby objects, aiding hazard anticipation. \item[(iv)] \textbf{Planning state:} Encodes driving context as $K$-dimensional binary vectors $\mathbf{Y}t$, derived through \emph{Yes}/\emph{No} queries. This representation helps prioritize actions and optimize routing. \item[(v)] \textbf{High-level planning and justification:} Decomposes driving decisions into meta-actions, which are mapped by a learnable encoder $E{A}$ into features $\mathbf{A}_t$. This modular design supports flexible, constraint-aware planning. \end{itemize}

Each QA task refines the vision-language model’s understanding of the scene, ensuring both lower-level perception and higher-level planning remain adaptive and interpretable.

We feed the planning state and meta-action features into the fast system, creating a human-like decision-making loop. Additionally, we introduce two prompts to enhance QA: (i) a visual prompt for human-like interpretation of scene elements, and (ii) a BEV prompt for a top-down perspective of spatial relationships.

\textbf{Visual Prompt:} In typical autonomous driving systems, waypoints generated by high-level planners are numerical outputs \cite{jiang2023vad, hu2023planning}. However, VLMs are not inherently designed to process numerical data in this context. Human decision-making in complex driving scenarios relies more on intuitive reasoning and visual cues than on direct numerical computation. To bridge this gap, we integrate trajectory visual prompts into our slow system planning. Specifically, we project the waypoints generated by the fast system planner onto the front-view camera, creating a visual representation of the trajectory, $\textbf{V}^{f}_t$. This visual approximation of the planned path facilitates human-like reasoning processes, enabling more intuitive evaluation and modification of decisions, which leads to more reliable and effective high-level plans.

\textbf{BEV Prompt:} To further enhance the system's spatial understanding, we introduce a BEV prompt. Based on the vehicle’s BEV coordinate system, this prompt provides a clear depiction of spatial relationships between the ego vehicle and surrounding agents, represented as $\textbf{B}_t$.

In summary, The slow system pipeline can be formulated as follows:
\begin{equation}
    \textbf{P}_t, \textbf{A}_t = \Phi(E(\textbf{V}_t^{f}), E(\textbf{B}_t))
\end{equation}

\subsubsection{VLM Tuning} \label{sec:vlm-tunning}

To adapt the VLM for QA-based planning, we combine an auto-labeled dataset with a reward-guided training scheme:

\begin{itemize} \item[(i)] \textbf{QA Dataset Generation:} We automatically annotate 3D detection boxes and tracked trajectories from the fast system, then leverage VLMs such as QwenVL~\cite{Qwen-VL} to produce descriptive QA pairs that align with the observed scene.

\item[(ii)] \textbf{Reward-Guided VLM Tuning:} Unlike standard LLM approaches reliant on pure auto-regressive learning, we incorporate both Maximum Likelihood Estimation (MLE) loss and a reward-guided regression loss. Inspired by, but distinct from, InstructGPT~\cite{mao2023gpt}, our method uses automatically generated guidance to replicate the planning state and high-level plans. Additionally, we integrate Proximal Policy Optimization (PPO)~\cite{schulman2017ppo} with masking to apply supervision at the token level, while treating the entire sequence as meaningful for regression. Concretely, we compute:
\begin{equation}
    \mathcal{L}_{\text{rvlm}} \;=\; \mathcal{F}_{\text{Reward}}\bigl(\mathbf{s}^{1:T_i}\bigr)\,\cdot\,\Phi\bigl(\mathbf{s}^{T_i}\,\big|\;\mathbf{s}^{1:T_i-1}\bigr),
\end{equation}
where $\mathbf{s}^{T_i}$ is the predicted token, $\mathcal{F}_{\text{Reward}}(\cdot)$ evaluates trajectories, and $\Phi(\cdot|\cdot)$ represents the policy. The final training objective combines the standard language loss and the reward-guided term:
\begin{equation}
\mathcal{L}_{\text{slow}} \;=\; \lambda_{\text{MLE}}\,\mathcal{L}_{\text{MLE}} \;+\; \lambda_{\text{rvlm}}\,\mathcal{L}_{\text{rvlm}}.
\end{equation}
\end{itemize}

\subsection{FAst and Slow Fusion Autonomous Driving}\label{fusion}

\textbf{Uncertainty Estimation:}
To effectively navigate dynamic and unpredictable environments, estimating uncertainty in waypoint predictions is essential, as it allows the system to adapt its decision-making based on prediction reliability. To handle outliers and model uncertainty in waypoint predictions, we employ a Laplace distribution:
\begin{equation}
    p(\text{R}\mid \Theta) = \prod_{t=1}^{T} \frac{1}{2b} \exp\left( -\frac{\|\mathbf{r}_t - \hat{\mathbf{\mu}}_t\|_1}{b} \right)
\end{equation}

Where \( \hat{\mathbf{\mu}}_t \) denotes the expectation of predicted reward at time \( t \), \( b \) is the scale parameter, $R$ is the reward, and \( \Theta \) represents the model parameters. The Laplace distribution's heavy tails and sharp peak make it robust to outliers and effective for uncertainty estimation in dynamic driving environments. The system uses the fast mode for planning when reward $R$ surpasses a set threshold with low uncertainty, and switches to the slow mode for detailed analysis in all other cases.

\textbf{Information Bottleneck:}
Driving environments often contain irrelevant or noisy information that does not contribute to planning. To address this, we apply the IB principle \cite{planKD} to distill the information relevant to decision-making.
Through interaction with environmental information, we derive the features of ego vehicle, denoted as \(z\). To align the learned planning-relevant representation \( z \) with the planning state \( y_t \), we employ the MLP layers that maps \( z \) to a one-dimensional vector \( y_i \). The knowledge distillation process minimizes the following objective:%MLP \( f_{\text{MLP}} \) 
\begin{equation}
\mathcal{L}_{\text{KD}} = \sum \log q_d(y_t | y_i) - \beta \, \text{KL}\left(q_e(y_i | z_{\text{current}}) \| p(z)\right)
\end{equation}
where \( q_d(y_t | y_i) \) is the probability distribution over the VLM-derived vector \( y_t \) given \( y_i \), and \( q_e(y_i | z_{\text{current}}) \) encodes query features from the current state. Here, \( p(z) \) is a prior distribution on \( z \), and \( \beta \) is a regularization parameter.

\textbf{High-level Action Guidance:}

To integrate high-level plans with the fast system, cross-attention is implemented between learnable embeddings $ E_{A} \in \mathbb{R}^{N_A \times d_A} $ and the ego token $ e_{\text{ego}} \in \mathbb{R}^{d_A} $. Specifically, the ego token $ E_{\text{ego}} $ queries $ E_{A} $ as key-value pairs. This process can be mathematically represented as follows:
\begin{equation}
    \text{Attention}(Q, K, V) = \text{softmax}\left(\frac{QK^T}{\sqrt{d_k}}\right)V
\end{equation}
where $ Q $, $ K $, and $ V $ represent the query, key, and value matrices respectively. The query matrix $ Q $ is derived from the ego token: $ Q = W_Q E_{\text{ego}} $; The key matrix $ K $ and value matrix $ V $ are derived from the learnable embeddings: $ K = W_K E_{A} $ and $ V = W_V E_{A} $, with $ W_Q \in \mathbb{R}^{d_k \times d_A} $, $ W_K \in \mathbb{R}^{d_k \times d_A} $, and $ W_V \in \mathbb{R}^{d_v \times d_A} $ being learnable weight matrices.

\section{Experiments}
In this section, we conduct experiments to address the following questions: (1) \textit{Does our feedback mechanism improve the planning performance of the fast E2E model?} (2) \textit{How does the uncertainty estimation meets the needs of handling complex driving scenarios?} (3) \textit{Do our information bottleneck and high-level plan instructions enhance the planning process?} (4) \textit{Does the VLM equipped with "visual and BEV prompts" provide a reasonable and transparent planning process?}

\subsection{Experimental Setup}

We evaluate \textbf{FASIONAD} on three leading autonomous driving benchmarks: \textbf{nuScenes}~\cite{nuscenes_benchmark}, \textbf{Town05 Short}~\cite{carla_benchmark}, and latest \textbf{Bench2Drive}~\cite{b2d}. Our evaluation encompasses both open-loop and closed-loop performance metrics. For open-loop assessment on nuScenes and Bench2Drive, we measure trajectory prediction accuracy against expert demonstrations using L2 distance and collision rate metrics. Specifically, we follow the default configuration of Bench2Drive, utilizing a base subset of 1,000 segments (950 for training and 50 for validation) with balanced scene and weather distributions. The implementation details are the same as for nuScenes, but we only trained the model for 6 episodes. The closed-loop evaluation on CARLA Town05 Short Benchmark measures Driving Score (DS)—calculated as the product of Route Completion (RC) and Infraction Score—and Route Completion itself. To ensure fair comparison, we implement a rule-based wrapper around our learning-based policy, following standard benchmark practices to minimize infractions during testing. Unless otherwise specified, experiments were conducted on a server equipped with 8 NVIDIA A100 GPUs.

\subsection{Main Results} 

\subsubsection{Open-loop Evaluation on nuScenes} We compare FASIONAD against traditional fast-system methods (e.g., VAD, GenAD) that rely purely on E2E trajectory prediction, as well as slow-system approaches (e.g., Agent-Driver) that leverage VLMs for decision-making. Additionally, we benchmark against dual-system frameworks, such as DriveVLM, which integrates vision-language reasoning into trajectory planning. As shown in Tab. \ref{table:nuscenes_comparison}, FASIONAD consistently outperforms all baseline models across different prediction horizons (1s, 2s, and 3s), achieving the lowest L2 trajectory error (0.28m on average) and the lowest collision rate (0.09\%). When compared to DriveVLM, our framework reduces the average L2 error by 9.6\% (from 0.31m to 0.28m) and further improves safety by reducing the collision rate from 0.10\% to 0.09\%. Notably, when paired with GenAD, it reduces the average L2 trajectory error by 24.2\% (from 0.91m to 0.69m) and the collision rate by 58.1\% (from 0.43\% to 0.18\%). Similarly, when using VAD-Base, FASIONAD achieves an 18.8\% improvement in L2 accuracy (from 1.22m to 0.99m) and a 49.1\% reduction in collision rate (from 0.53\% to 0.27\%). This improvement highlights the effectiveness of the adaptive switching mechanism in improving trajectory accuracy and safety in different slow systems.

\begin{figure}[ht]
    \centering
    % \vspace{-10pt}
    \includegraphics[width=\linewidth]{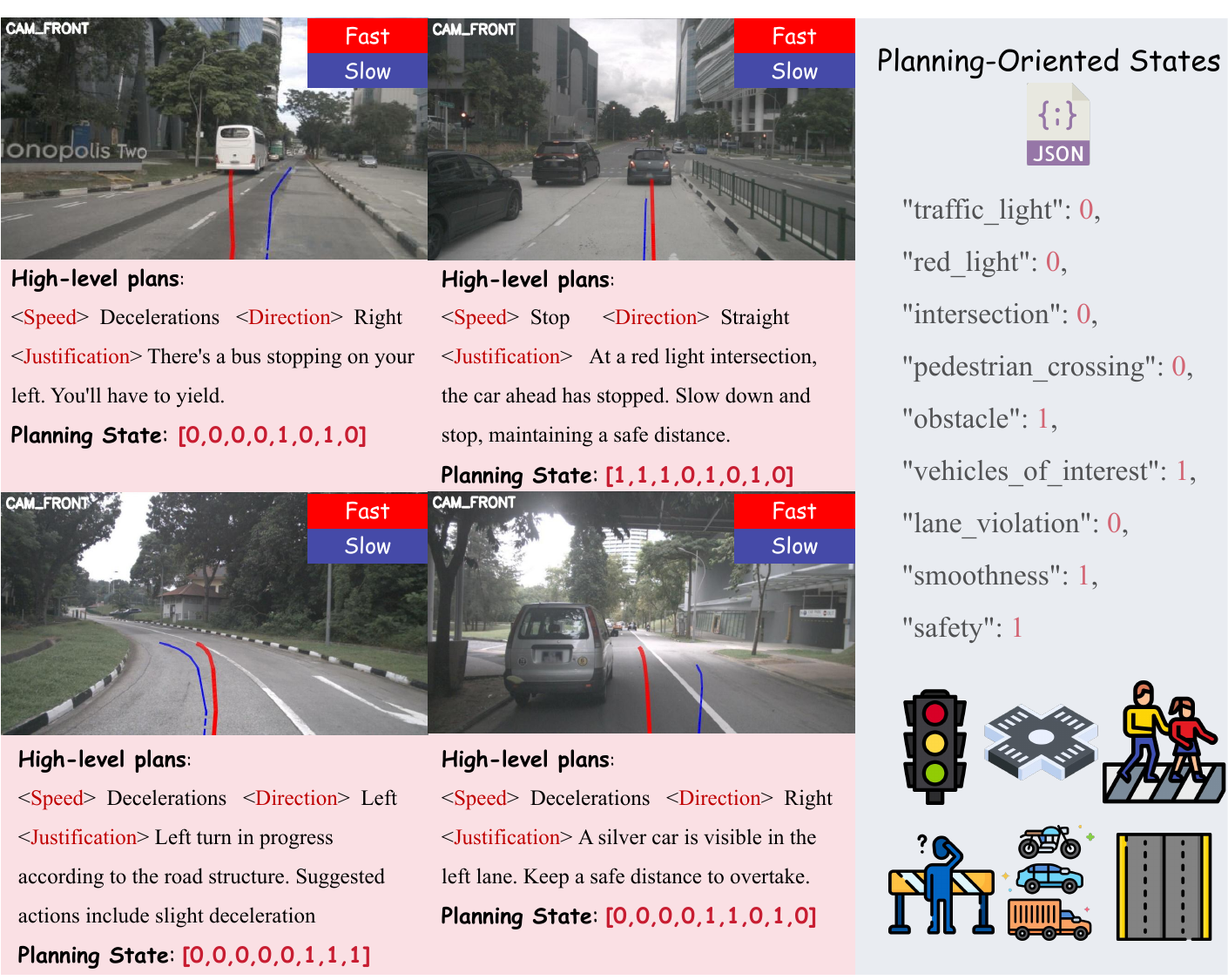}
    \vspace{-20pt}
    \caption{Example scenarios demonstrating FASIONAD's adaptive feedback framework in various driving environments. Each scene shows different navigation challenges, including obstacles, lane adjustments, and turns. The proposed system provides suggested driving operations and ensures safe, smooth trajectories with minimal abrupt maneuvers, enhancing safety in complex situations.}
    \vspace{-10pt}
    \label{fig:adaptivefeedbackl}
\end{figure}

To better illustrate the effectiveness of the switching mechanism, Fig. \ref{fig:adaptivefeedbackl} gives examples where FASIONAD successfully adapts to complex planning tasks.
When approaching an intersection, the system dynamically adjusts its trajectory. On highways, it assesses traffic conditions and selectively activates the slow system for safe and efficient lane changes. At signalized intersections, FASIONAD accurately interprets traffic lights and obstacles, ensuring timely stops and maintaining safe distances. During overtaking, if the fast system's generated trajectory is deemed unsafe, FASIONAD maintains a safe distance, using deceleration instructions to ensure a secure overtake.
By integrating a structured planning state and high-level plans with adaptive feedback, the system improves decision-making interpretability and planning safety.

\begin{table*}[htbp]
\centering
\caption{Open-loop planning performance on the nuScenes validation dataset}

\label{table:nuscenes_comparison}
\small
\begin{tabular}{lccccccccccc}
\toprule
\textbf{Method} & \textbf{Input} &\textbf{System}& \multicolumn{4}{c}{\textbf{L2 (m) $\downarrow$}} & \multicolumn{4}{c}{\textbf{Collision Rate (\%) $\downarrow$}} & \textbf{FPS} \\
\cmidrule(lr){4-7} \cmidrule(lr){8-11}
& & & \textit{1s} & \textit{2s} & \textit{3s} & \textit{Avg} & \textit{1s} & \textit{2s} & \textit{3s} & \textit{Avg} & \\
\midrule
IL~\cite{ratliff2006maximum} & LiDAR & - & 0.44 & 1.15 & 2.47 & 1.35 & 0.08 & 0.27 & 1.95 & 0.77 & - \\
NMP~\cite{NMP} & LiDAR & Fast.  & 0.53 & 1.25 & 2.67 & 1.48 & 0.04 & 0.12 & 0.87 & 0.34 & - \\
FF~\cite{FF} & LiDAR & Fast. & 0.55 & 1.20 & 2.54 & 1.43 & 0.06 & 0.17 & 1.07 & 0.43 & - \\
EO~\cite{EO} & LiDAR & Fast. & 0.67 & 1.36 & 2.78 & 1.60 & 0.04 & 0.09 & 0.88 & 0.33 & - \\
ST-P3 ~\cite{hu2022stp3} & Camera & Fast. & 1.33 & 2.11 & 2.90 & 2.11 & 0.23 & 0.62 & 1.27 & 0.71 & 1.6 \\
OccNet \cite{OCCNet} & Camera & Fast. & 1.29 & 2.13 & 2.99 & 2.14 & 0.21 & 0.59 & 1.37 & 0.72 & 2.6 \\
UniAD ~\cite{hu2023planning} & Camera & Fast. & 0.48 & 0.96 & 1.65 & 1.03 & 0.05 & 0.17 & 0.71 & 0.31 & 1.8 \\
% \textbf{UniAD w/ Ours} & Camera & 0.48 & 0.96 & 1.65 & 1.03 & 0.05 & 0.17 & 0.71 & 0.31 & 1.8 \\
\midrule\midrule
Agent-Driver~\cite{mao2024languageagent} & Camera & Slow. &  0.22 & 0.65 &  1.34 &  0.74 &  0.02 &  0.13 &  0.48 &  0.21 & None† \\
DriveVLM* \cite{tian2024drivevlm} & Camera & Dual. & 0.15 & 0.29 & 0.48 & 0.31 & 0.05 & 0.08 & 0.17 & 0.10 & 6.7† \\
\cellcolor{waymolgray}\textbf{FASIONAD*} w/ GenAD & \cellcolor{waymolgray}Camera & \cellcolor{waymolgray}Dual. & \cellcolor{waymolgray}\textcolor{waymoblue}{\textbf{0.13}} & \cellcolor{waymolgray}\textcolor{waymoblue}{\textbf{0.26}} & \cellcolor{waymolgray}\textcolor{waymoblue}{\textbf{0.45}} & \cellcolor{waymolgray}\textcolor{waymoblue}{\textbf{0.28}} & \cellcolor{waymolgray}\textcolor{waymoblue}{\textbf{0.05}} & \cellcolor{waymolgray}\textcolor{waymoblue}{\textbf{0.08}} & \cellcolor{waymolgray}\textcolor{waymoblue}{\textbf{0.15}} & \cellcolor{waymolgray}\textcolor{waymoblue}{\textbf{0.09}} & \cellcolor{waymolgray}\textcolor{waymoblue}{\textbf{7.1}}† \\
\midrule
VAD-Tiny ~\cite{jiang2023vad} & Camera & Fast. & 0.60 & 1.23 & 2.06 & 1.30 & 0.31 & 0.53 & 1.33 & 0.72 & 6.9† \\
VAD-Base ~\cite{jiang2023vad} & Camera & Fast. & 0.54 & 1.15 & 1.98 & 1.22 & 0.04 & 0.39 & 1.17 & 0.53 & 3.6† \\
\cellcolor{waymolgray}\textbf{FASIONAD} w/ VAD-Base & \cellcolor{waymolgray}Camera & \cellcolor{waymolgray}Dual. & \cellcolor{waymolgray}0.41 & \cellcolor{waymolgray}0.95 & \cellcolor{waymolgray}1.62 & \cellcolor{waymolgray}0.99 & \cellcolor{waymolgray}0.03 & \cellcolor{waymolgray}0.15 &\cellcolor{waymolgray} 0.62 & \cellcolor{waymolgray}0.27 & \cellcolor{waymolgray}3.9† \\
GenAD \cite{GenAD} & Camera & Fast. & 0.25 & 0.68 & 1.30 & 0.74 & 0.05 & 0.17 & 0.53 & 0.25 & 6.7† \\

\cellcolor{waymolgray}\textbf{FASIONAD} w/ GenAD & \cellcolor{waymolgray}Camera &\cellcolor{waymolgray} Dual. & \cellcolor{waymolgray}\textcolor{waymoblue}{\textbf{0.19}} & \cellcolor{waymolgray}\textcolor{waymoblue}{\textbf{0.62}} & \cellcolor{waymolgray}\textcolor{waymoblue}{\textbf{1.25}} & \cellcolor{waymolgray}\textcolor{waymoblue}{\textbf{0.69}} & \cellcolor{waymolgray}\textcolor{waymoblue}{\textbf{0.02}} & \cellcolor{waymolgray}\textcolor{waymoblue}{\textbf{0.09}} & \cellcolor{waymolgray}\textcolor{waymoblue}{\textbf{0.44}} & \cellcolor{waymolgray}\textcolor{waymoblue}{\textbf{0.18}} & \cellcolor{waymolgray}\textcolor{waymoblue}{\textbf{7.0}}† \\

\bottomrule
\end{tabular}
\begin{flushleft}
\small
\textbf{Note:} * denotes using ego status features as input. † represents that the metrics are computed with an average of all the predicted frames. † denotes FPS measured in the same environment on our machine with a single RTX 3090 GPU. 
\end{flushleft}
\end{table*}

\subsubsection{Open-loop evaluation on Bench2Drive}

Tab. \ref{tab:comparison} shows comparison results with several well-established E2E methods. FASIONAD achieves an L2 error of 0.82m and a collision rate of 0.12\%, demonstrating notable improvements over VAD (0.91m, 0.19\%). While UniAD-Base achieves a slightly lower L2 error (0.73m), its reliance on deterministic trajectory generation without explicit uncertainty modeling may lead to increased safety risks in real-world deployment. Compared to AD-MLP, which has a much higher L2 error of 3.64m, our method benefits from its adaptive feedback mechanism, improving both accuracy and safety. These results highlight the effectiveness of FASIONAD’s feedback-driven adaptation, where high-level vision-language reasoning complements fast trajectory generation, leading to more precise and safer predictions across diverse traffic scenarios.
 \begin{table}[t]
    \centering
    \caption{Comparison of methods based on Bench2Drive benchmark Open-loop Evaluation}
    \begin{tabular}{lcc}
        \toprule
        \textbf{Method} & \textbf{Avg. L2 ↓} & \textbf{Avg. C.R. ↓} \\
        \midrule
        AD-MLP~\cite{zhai2023rethinking} & 3.64 & - \\
        UniAD-Tiny~\cite{hu2023planning} & 0.80 & - \\
        UniAD-Base~\cite{hu2023planning} & \textbf{0.73} & - \\
        VAD~\cite{jiang2023vad} & 0.91 & 0.19 \\
        \textbf{FASIONAD} w/ VAD-Base & 0.82 & \textbf{0.12} \\
        % \midrule
        % TCP*~\cite{wu2022trajectory} & 1.70 & - \\
        % TCP-ctrl* & - & - \\
        % TCP-traj* & 1.70 & - \\
        % TCP-traj w/o distillation & 1.96 & - \\
        % ThinkTwice*~\cite{jia2023think} & \textbf{0.95} & - \\
        % DriveAdapter*~\cite{jia2023driveadapter} & 1.01 & - \\
        \bottomrule
    \end{tabular}
    \label{tab:comparison}
\begin{flushleft}
\small
\textbf{Note:} Avg. L2 is calculated similarly to the UniAD. %averaged over the predictions in 2 seconds under 2Hz,  
\end{flushleft}
\vspace{-10pt}
\end{table}

\subsubsection{Closed-loop evaluation on CARLA}
To validate FASIONAD's driving skills in closed-loop evaluations, we compare our proposed FASIONAD with a variety of published algorithms. Tab. \ref{tab:driving_methods_comparison} presents a comparative analysis against state-of-the-art E2E autonomous driving models such as multi-modal based Transfuser \cite{chitta2022transfuser}, query-based VAD \cite{jiang2023vad}, and LLM-based methods Agent-Driver \cite{mao2024languageagent}. FASIONAD achieves the highest  DS (64.83\%) and RC (89.04\%), surpassing prior methods in both driving stability and route-following accuracy. These results demonstrate that our approach not only improves planning accuracy in open-loop settings but also enhances overall driving performance in interactive scenarios.

\definecolor{bestgreen}{rgb}{0.0, 0.6, 0.0}
\begin{table}[t]
\centering
%\vspace{-10pt}
\caption{Closed-loop evaluation on Town05 Short benchmark}
\footnotesize  
\setlength{\tabcolsep}{5pt} 
\label{tab:driving_methods_comparison}
\begin{tabular}{lccc}
\hline
\textbf{Methods} & \textbf{Modality} & \textbf{DS (\%) $\uparrow$} & \textbf{RC (\%) $\uparrow$} \\ \hline
CILRS ~\cite{codevilla2019exploring} & C &7.47 & 13.40 \\ 
LBC  ~\cite{cui2021lookout} & C &30.97 & 55.01 \\ 
Transfuser ~\cite{chitta2022transfuser} & C+L &54.52 & 78.41 \\ 
ST-P3 ~\cite{hu2022stp3} & C &55.14 & 86.74 \\ 
VAD ~\cite{jiang2023vad} & C &64.29 & 87.26 \\ 
\midrule
Agent-Driver ~\cite{mao2024languageagent}& C &64.31 & 87.31 \\ 
\cellcolor{waymolgray}\textbf{FASIONAD} w/ GenAD & \cellcolor{waymolgray}C & \cellcolor{waymolgray}\textcolor{waymoblue}{\textbf{64.83}} & \cellcolor{waymolgray}\textcolor{waymoblue}{\textbf{89.04}} \\ \hline
\end{tabular}
\end{table}

\subsubsection{Explainability and reliability in planning states and high-level plans}
Since FASIONAD integrates VLMs to enhance trajectory planning, we conduct experiments following the RAG-Driver \cite{yuan2024rag} setup to quantitatively analyze different models' performance on planning state recognition, high-level action prediction, and explanation quality (BLEU-4, CIDEr, METEOR), as shown in Tab. \ref{table:vlm}. Results show that task-specific prompts significantly improve all models, with Video-LLaVA achieving the highest accuracy (55.74\% planning state, 62.85\% high-level action) and best explainability (25.34 BLEU-4, 50.48 METEOR). While InternVL and QwenVL also perform well, their improvements are less pronounced. The substantial performance gap between standard and task-specific prompts highlights the importance of structured input, aligning with FASIONAD’s approach of integrating VLM-guided reasoning to enhance planning accuracy and interpretability.

\begin{table}[t]
    \centering
    \caption{Comparison of VLMs in Planning-Oriented Tasks}
    \begin{tabular}{l@{\hspace{5pt}}c@{\hspace{5pt}}c@{\hspace{5pt}}c@{\hspace{5pt}}c@{\hspace{5pt}}c}
        \toprule
        \textbf{Method} & \textbf{Plan. S.} & \textbf{High. A.} & \textbf{BLEU.}$\uparrow$  & \textbf{CID.}$\uparrow$  & \textbf{MET.}$\uparrow$  \\
        & \textbf{Acc}$\uparrow$ & \textbf{Acc}$\uparrow$  & & & \\
        \midrule
        % \midrule
        QwenVL \cite{Qwen-VL} & 15.13 & 9.75 & 9.45 & 6.22 & 32.11 \\
        InternVL \cite{chen2024internvl} & \textbf{17.92} & \textbf{10.18} & \textbf{12.62} & \textbf{7.32} & \textbf{35.68} \\
        Video-LLaVA \cite{lin2023videollava} & 8.46 & 8.14 & 8.85 & 4.19 & 31.92 \\
        \midrule
        % \midrule
        QwenVL \cite{Qwen-VL}\textdagger & 52.33 & 61.87 & 24.77 & 20.13 & 48.25 \\
        InternVL \cite{chen2024internvl}\textdagger & \textbf{56.81} & 62.45 & 23.41 & \textbf{20.84} & 48.19 \\
        Video-LLaVA \cite{lin2023videollava}\textdagger & 55.74 & \textbf{62.85} & \textbf{25.34} & 19.53 & \textbf{50.48} \\
        \bottomrule
    \end{tabular}
    \label{table:vlm}
\begin{flushleft}
\small
\textbf{Note:} The \textdagger denotes configurations equipped with task-specific prompts using our proposed planning-oriented QAs. Plan. S. and High. A. relatively denote planning states and high-level actions.
\end{flushleft}
\vspace{-10pt}
\end{table}

\begin{table}[t]
\caption{Ablation Study of IB and HA}
\footnotesize  % 使用更小的字体
\setlength{\tabcolsep}{4pt} % 减小列间距
\centering
\small 
\begin{tabular}{cc|cccc|cccc}
\toprule
\multicolumn{2}{c|}{\textbf{ Setting}} & \multicolumn{4}{c|}{\textbf{L2 (m) $\downarrow$}} & \multicolumn{4}{c}{\textbf{Collision Rate (\%) $\downarrow$}} \\
\cmidrule(lr){1-2} \cmidrule(lr){3-6} \cmidrule(lr){7-10}
\textbf{IB} & \textbf{HA} & \textbf{1s} & \textbf{2s} & \textbf{3s} & \cellcolor{waymolgray}\textbf{Avg.} & \textbf{1s} & \textbf{2s} & \textbf{3s} & \cellcolor{waymolgray}\textbf{Avg.} \\
\midrule
\cellcolor{waymolgray}\cmark & \xmark & 0.23 & 0.66 & 1.34 & \cellcolor{waymolgray}0.74 & 0.03 & 0.12 & 0.47 & \cellcolor{waymolgray}0.21 \\
\xmark & \cellcolor{waymolgray}\cmark & 0.24 & 0.68 & 1.37 & \cellcolor{waymolgray}0.77 & \textbf{0.02} & 0.10 & 0.45 & \cellcolor{waymolgray}0.19 \\
\cellcolor{waymolgray}\cmark & \cellcolor{waymolgray}\cmark & \textbf{0.19} & \textbf{0.62} & \textbf{1.25} & \cellcolor{waymolgray}\textbf{0.69} & \textbf{0.02} & \textbf{0.09} & \textbf{0.44} & \cellcolor{waymolgray}\textbf{0.18} \\

\bottomrule
\end{tabular}
\label{tab:info_bottleneck_high_level_action}
\end{table}

\begin{table}[htbp]
\vspace{-8pt}
\caption{Ablation study of uncertainty module}
\vspace{-10pt}
\footnotesize  % 使用更小的字体
\setlength{\tabcolsep}{3pt} % 减小列间距
\centering
\small 
\resizebox{\linewidth}{!}{
    \begin{tabular}{cc|cccc|cccc|cc}
        \toprule
        \multicolumn{2}{c|}{\textbf{Setting}} & \multicolumn{4}{c|}{\textbf{L2 (m) $\downarrow$}} & \multicolumn{4}{c|}{\textbf{Collision Rate(\%) $\downarrow$}} & \multicolumn{2}{c}{\textbf{VLM Trigger Rate}}\\
        \cmidrule(lr){1-2} \cmidrule(lr){3-6} \cmidrule(lr){7-10} \cmidrule(lr){11-12} 
        \textbf{Asyn.} & \textbf{Uncer.} & \textbf{1s} & \textbf{2s} & \textbf{3s} & \cellcolor{waymolgray}\textbf{Avg.} & \textbf{1s} & \textbf{2s} & \textbf{3s} & \cellcolor{waymolgray}\textbf{Avg.} & \textbf{Fast} & \textbf{Slow$\downarrow$}\\
        \midrule
        \cellcolor{waymolgray}\cmark & \xmark & \textbf{0.17} & \textbf{0.60} & \textbf{1.22} & \cellcolor{waymolgray}0.66 & 0.02 & 0.08 & 0.45 & \cellcolor{waymolgray}0.18 & 76.92\% & 23.08\% \\
        \xmark & \cellcolor{waymolgray}\cmark & 0.19 & 0.62 & 1.25 & \cellcolor{waymolgray}0.69 & \textbf{0.02} & 0.09 & \textbf{0.44} & \cellcolor{waymolgray}\textbf{0.18} & 91.26\% & \textbf{8.74\%}(\textcolor{waymoblue}{\textbf{$\downarrow$62.13\%}})  \\
        \bottomrule
    \end{tabular}
}
% \vspace{-10pt}
\label{tab:uncertainty}
\end{table}
\subsection{Ablation Study}
In this section, we implement the models without ego-state to purely evaluate the components. The fast E2E model used is GenAD  \cite{GenAD}.
\subsubsection{Modular designs} 

Our ablation study demonstrates the complementary benefits of the IB and HA components (Tab. \ref{tab:info_bottleneck_high_level_action}). The full model incorporating both components achieved the best performance (L2: 0.69m, collision rate: 0.18\%). Using either component alone led to decreased performance - IB-only (L2: 0.74m,
collision rate: 0.21\%) and HA-only (L2: 0.77m, collision rate: 0.19\%) - highlighting their synergistic relationship in improving prediction accuracy through effective information filtering and high-level planning. To assess the impact of the uncertainty estimation mechanism, we conduct an ablation study comparing two setups in Tab.\ref{tab:uncertainty}: (1) triggering the fast-slow systems asynchronously, and (2) incorporating uncertainty estimation. With the uncertainty switch, planning performance remains stable while reducing computational load. Specifically, the VLM trigger rate decreases by 62.13\% compared to asynchronous methods (e.g., DriveVLM\cite{tian2024drivevlm}).

\begin{table}[ht]
\caption{Validation of VLM Prompt Strategies}
% \vspace{-10pt}
\footnotesize 
\setlength{\tabcolsep}{4pt} 
\centering
\small 
\begin{tabular}{c|cccc|cccc}
\toprule
\textbf{Setting} & \multicolumn{4}{c|}{\textbf{L2 (m) $\downarrow$}} & \multicolumn{4}{c}{\textbf{Collision Rate (\%) $\downarrow$}} \\
\cmidrule(lr){2-5} \cmidrule(lr){6-9}
& \textbf{1s} & \textbf{2s} & \textbf{3s} & \cellcolor{waymolgray}\textbf{Avg.} & \textbf{1s} & \textbf{2s} & \textbf{3s} & \cellcolor{waymolgray}\textbf{Avg.} \\
\midrule
Simple.P& 0.31 & 0.71 & 1.38 & \cellcolor{waymolgray}0.80 & 0.05 & 0.16 & 0.74 & \cellcolor{waymolgray}0.32 \\
BEV.P  & 0.29 & 0.70 & 1.36 & \cellcolor{waymolgray}0.79 & 0.04 & 0.14 & 0.65 & \cellcolor{waymolgray}0.24 \\
Visual.P  & 0.24 & 0.67 & 1.30 & \cellcolor{waymolgray}0.74 & \textbf{0.02} & 0.11 & 0.48 & \cellcolor{waymolgray}0.20 \\
Full.P & \textbf{0.19} & \textbf{0.62} & \textbf{1.25} & \cellcolor{waymolgray}\textbf{0.69} & \textbf{0.02} & \textbf{0.09} & \textbf{0.44} & \cellcolor{waymolgray}\textbf{0.18} \\
\bottomrule
\end{tabular}
\label{tab:vlm_io_effect}
\begin{flushleft}
\small
\textbf{Note:} In the setting, "P" denotes a prompt (e.g., BEV.P indicates a BEV prompt).).
\end{flushleft}
\vspace{-10pt}
\end{table}

\subsubsection{VLM prompt strategy} 
Our ablation study on VLM prompt strategies revealed the significant impact of prompt design (Tab.~\ref{tab:vlm_io_effect}). The Full.P configuration, featuring comprehensive prompt instructions, achieved the best results with an L2 distance of 0.69 meters and 0.18\% collision rate. Performance gradually declined with simpler prompting approaches: Visual.P (0.74m, 0.20\%), BEV.P (0.79m, 0.24\%), and Simple.P (0.80m, 0.32\%). These results demonstrate that detailed, well-structured prompts are crucial for maximizing VLM's predictive capabilities.

\section{Conclusion and Future Work}
In this paper, we presented \textbf{FASIONAD}, a dual‐system autonomous driving framework that unifies an E2E fast planner with a VLM‐based slow system to address high‐uncertainty or complex traffic scenarios. Our Uncertainty Estimation selectively triggers the slow system only when deeper reasoning is required, while Information Bottleneck and High‐Level Action Guidance serve as targeted feedback loops to enhance the fast system’s planning efficiency. Additionally, our integration of visual and BEV prompts, combined with reward‐guided VLM training, provides interpretability and robustness. Experiments on nuScenes, Town05 Short, and Bench2Drive confirm that FASIONAD not only improves trajectory accuracy but also substantially reduces collision rates, all while maintaining computational efficiency. In future work, we plan to extend FASIONAD to unstructured or rural settings and explore additional sensor modalities to further expand its robustness.

% %%%%%%%%%%%%%%%%%%%%%%%%%%%%%%%%%%%%%%%%%%%%%%%%%%%%%%%%%%%%%%%%%%%%%%%%%%%%%%%%

% References are important to the reader; therefore, each citation must be complete and correct. If at all possible, references should be commonly available publications.

\bibliographystyle{IEEEtran}
\bibliography{references}

\end{document}